\documentclass[sigconf]{acmart}
\usepackage{graphicx}   
\usepackage{booktabs}   
\usepackage{multirow}   
\usepackage{makecell}   
\usepackage{colortbl}   
\usepackage{balance}
\AtBeginDocument{%
  }

\copyrightyear{2026}
\acmYear{2026}
\setcopyright{cc}
\setcctype{by}
\acmConference[MM '26]{Proceedings of the 34th ACM International Conference on Multimedia}{November 10--14, 2026}{Rio de Janeiro, Brazil}
\acmBooktitle{Proceedings of the 34th ACM International Conference on Multimedia (MM '26), November 10--14, 2026, Rio de Janeiro, Brazil}
\acmDOI{10.1145/3767308.3836149}
\acmISBN{979-8-4007-2213-4/2026/11}

\begin{document}

\title{DGNet: Dual-knowledge Guided Network for Infrared Small Target Detection}

\author{Chenglong Yu}
\orcid{0009-0004-7729-1616}
\affiliation{
  \institution{\normalsize Shandong University}
  \city{Jinan}
  \country{China}
  }
\email{yucl@mail.sdu.edu.cn}
\author{Mingzhu Xu}
\orcid{0000-0002-1492-0970}
\authornote{Corresponding authors: Mingzhu Xu and Liqiang Nie.}
\affiliation{
  \institution{\normalsize Shandong University}
  \city{Jinan}
  \country{China}
  }
\email{xumingzhu@sdu.edu.cn}

\author{Jing Wang}
\orcid{0009-0003-1107-2528}
\affiliation{
  \institution{\normalsize Shandong University}
  \city{Jinan}
  \country{China}
  }
\email{202415291@mail.sdu.edu.cn}

\author{Tongtong Wang}
\orcid{0009-0003-2421-1363}
\affiliation{%
  \institution{Shandong University}
  \city{Jinan}
  \country{China}
}
\email{wangtongtong0116@163.com}

\author{Pingping Miao}
\orcid{0009-0001-1531-1106}
\affiliation{
  \institution{\normalsize Shandong University}
  \city{Jinan}
  \country{China}
  }
\email{miaopp@mail.sdu.edu.cn}

\author{Liqiang Nie}
\orcid{0000-0003-1476-0273}
\authornotemark[1]
\affiliation{
  \institution{\normalsize Harbin Institute of Technology, Shenzhen}
  \city{Shenzhen}
  \country{China}
  }
\email{nieliqiang@gmail.com}

\renewcommand{\shortauthors}{Chenglong Yu et al.}

\begin{abstract}
InfRared Small Target Detection (IRSTD) is a prominent and challenging task in computer vision. In recent years, text-guided methods have significantly improved detection performance. However, they still suffer from two key limitations. First, a single text description simultaneously modeling both background and target leads to semantic entanglement, which contradicts the objective of background suppression and target enhancement. Second, reliance on image-specific textual prompts (requiring additional external models such as CLIP during inference) results in deployment constraints. To address these issues, we propose a novel Dual-knowledge Guided Network (DGNet) based on multiple generalizable texts. Specifically, we design a Prior-knowledge Wavelet Modulation (PWM) module, which leverages dual textual priors that separately characterize large-scale backgrounds and sparse targets to effectively disentangle and modulate entangled semantics in the frequency domain. Furthermore, we introduce a Consensus-knowledge Directional Alignment (CDA) loss, which models the initial state and the ideal target across samples as `complex background' and `bright target', respectively, thereby constructing a clear and unified directional optimization trajectory for the model. Extensive experiments on three public datasets demonstrate the superior performance of DGNet and the effectiveness of each component. The source code is available at \href{https://github.com/iLearn-Lab/MM26-DGNet}{https://github.com/iLearn-Lab/MM26-DGNet}.
\end{abstract}







\maketitle
\section{Introduction}
\label{sec:intro}
InfRared Small Target Detection (IRSTD) aims to accurately localize tiny targets with low signal-to-noise ratios, playing an irreplaceable role in both civil and military applications~\cite{surveillance,Empowering,wcdmfnet,hdnet,mpcnet,essnet,stpinnet}. However, due to long-distance imaging and thermal radiation characteristics, infrared small targets typically occupy only a few pixels in the image and lack distinct color, shape, or texture features~\cite{mmlnet,tip_2,HFCNet2024,moving_aaai_2025}. Moreover, complex background clutter in real-world scenarios makes robust target segmentation highly challenging\cite{l2sknet,xmz1,xmz2,sgbd,xmz4,denet,cuixueliang}.

Early traditional methods, including filter-based methods \cite{filter_2,filter_3,filter_4,filter_6,top-hat}, local contrast-based methods \cite{contrast_1,contrast_2,contrast_3,contrast_4}, and low-rank representation methods \cite{low_rank_1,low_rank_2,low_rank_3,low_rank_4,non_convex_rank}, have explored and partially alleviated the IRSTD problem. Deep learning (DL)-based methods~\cite{dnanet,isnet,semisupervised,mopkl-tgrs,alcnet,pknet,acmnet,sctransnet,pbt,rkformer,abc,dconet,drtenet,fscfnet} have significantly improved performance by learning hierarchical visual features. However, purely visual IRSTD methods rely solely on single-modal information, making it difficult for the model to extract discriminative features and leading to false alarms.

\begin{figure}[t]
    \centering
    \includegraphics[width=1.0\linewidth]{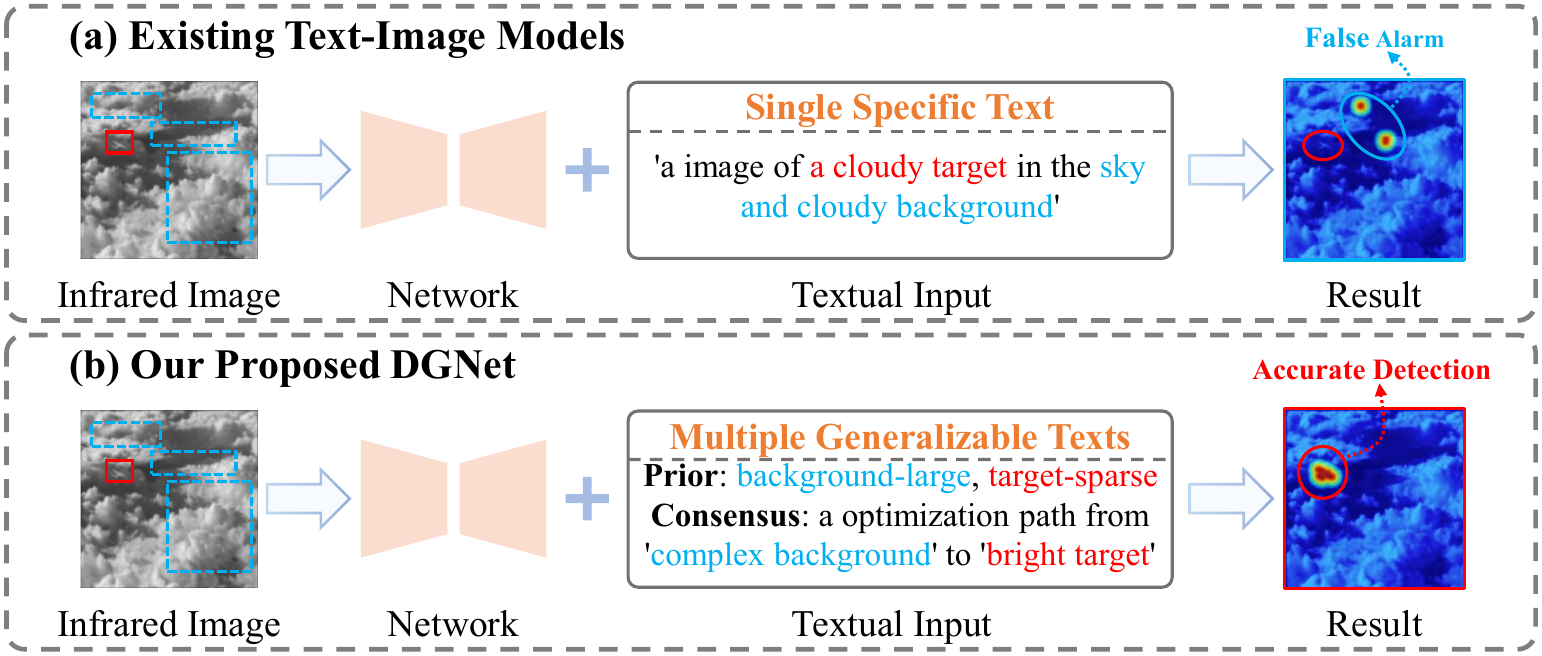}
    \vspace{-13pt}
    \caption{ Visual comparison of (a) Existing Text-Image Models and (b) Our Proposed DGNet on cluttered infrared image.}
    \vspace{-16pt}
    \Description[Visual comparison]{Comparison of IRSTD results on complex scenes. Pure visual models tend to be affected by background clutter and noise, leading to false alarms. Text–image fusion models relying on specific descriptions may overemphasize background semantics, resulting in missed targets. In contrast, the proposed DGNet leverages fixed prior and consensus knowledge to accurately highlight targets while effectively suppressing background interference.}
    \label{fig:Intro}
\end{figure}

In recent years, some pioneering works~\cite{text-irstd,saist} have introduced the textual modality as an auxiliary to the visual modality, a strategy akin to the semantic guidance explored in multimodal learning~\cite{xmz3,xmz5,xmz6}, significantly improving the accuracy of small target detection. However, as illustrated in Fig.~\ref{fig:Intro}(a), these methods typically adopt a single specific text as guidance. This paradigm suffers from two key limitations: \textbf{1) Semantic entanglement caused by a single text.} Infrared images are structurally composed of large, smooth backgrounds and small, sparse targets. Existing textual descriptions often jointly characterize both objects, such as `sky target' (sparse small targets) and `sky and cloudy background' (large-scale background) in Fig.~\ref{fig:Intro}(a). Such a single textual prompt encourages the network to process background and target information simultaneously, resulting in semantic entanglement that is difficult to disentangle. This contradicts the fundamental objective of IRSTD, which requires effective background suppression and target enhancement. Therefore, designing multiple texts that guide the model to handle these entangled semantics separately becomes the first challenge.
\textbf{2) Deployment limitations caused by image-specific texts.} Existing text-image models typically rely on image-specific textual prompts, where a dedicated description is generated for each image during both training and inference. However, this paradigm requires external models (e.g., CLIP~\cite{clip}) during inference, leading to increased computational overhead and significantly limiting practical deployment. Therefore, designing generalizable texts that enable the model to learn consensus knowledge across samples becomes the second challenge.

To address these challenges, we propose a novel Dual-knowledge Guided Network (DGNet). As shown in Fig.~\ref{fig:Intro}(b), unlike conventional methods that rely on a single image-specific text, DGNet leverages multiple generalizable texts, constructed from both prior knowledge and consensus knowledge. Specifically, in the feature extraction stage, we design a Prior-knowledge Wavelet Modulation (PWM) module, which utilizes the dual textual prior of separating large-scale backgrounds and sparse small targets to achieve decoupled modulation of the two entangled semantics in the frequency domain. In the optimization stage, we propose a Consensus-knowledge Directed Alignment (CDA) loss, which models the initial state of all samples as `complex background' and the ideal optimization endpoint as `bright targets', forming cross-sample consensus knowledge. Guided by this consensus, the CDA loss establishes a directed optimization path from `complex background' to `bright targets', providing a unified trajectory for model learning.

In summary, the main contributions of this paper are as follows:
\begin{itemize}
    \item We identify the limitations of existing text-image models, including semantic entanglement caused by single specific text guidance and deployment constraints introduced by image-specific texts. Based on this, we propose a novel DGNet driven by multiple generalizable texts.
    \item We design a novel PWM module leveraging dual textual priors to decouple entangled semantics in the frequency domain. Furthermore, we propose a new CDA loss incorporating consensus knowledge to guide the model's optimization path from `complex background' to `bright targets'.
    \item Extensive ablation studies and comparative experiments on three public datasets demonstrate the superiority of DGNet and the effectiveness of its key components.
\end{itemize}

\section{Related Work}
\label{sec:related}
\subsection{Infrared Small Target Detection}
IRSTD methods have evolved significantly over the years. 
Traditional approaches can be broadly categorized into three types. Filter-based methods \cite{filter_2,filter_3,filter_4} rely on handcrafted filters but struggle in complex scenarios. Local contrast-based methods \cite{contrast_1,contrast_2,contrast_3,contrast_4} enhance target saliency via surrounding comparisons but often cause false alarms. Low-rank methods \cite{low_rank_1,low_rank_2,low_rank_3,low_rank_4} perform well in smooth scenes but tend to miss targets in cluttered backgrounds. Overall, the heavy reliance on handcrafted priors limits their robustness in complex scenarios.
In contrast, DL-based methods \cite{ismallnet,ristdnet,ascnet} adopt a data-driven paradigm to learn target features and have achieved significant progress in IRSTD. For example, HDNet \cite{hdnet} introduces a hybrid-domain framework combining spatial multiscale atrous contrast and dynamic high-pass filtering to enhance small-target detection and suppress background.
IRPNet \cite{irpnet} introduce rich RGB knowledge into IRSTD, enhancing the model's representation capability.
DRPCA-Net \cite{drpcanet} integrates sparse representation priors into a learnable architecture, enabling accurate estimation of low-rank features.
Despite existing IRSTD methods have achieved significant progress, purely visual models struggle to extract discriminative features between targets and backgrounds, limiting detection performance.
To address the limitations of single-modality methods, recent studies have explored text-guided IRSTD. Benefiting from the strong cross-modal semantic modeling capability of CLIP~\cite{clip}, as well as semantic interaction~\cite{lzx1,lzx2,lzx7}, semantic-guided visual learning~\cite{zhanghaoyu2,liumeng,zhanghaoyu1}, and robust representation learning~\cite{lzx3,lzx4,lzx5,lzx6} explored in related tasks, methods such as Text-IRSTD \cite{text-irstd} and SAIST \cite{saist} incorporate textual descriptions to enhance detection performance.

However, these methods typically rely on a single textual prompt, which often jointly characterizes both backgrounds and targets. This unified textual modeling confounds heterogeneous semantics, forcing the network to process background and target information simultaneously. As a result, existing methods lack the ability to explicitly disentangle these entangled semantics under single-text guidance. To address this issue, we design the PWM module, which leverages dual textual priors that separately characterize large-scale backgrounds and sparse targets to effectively disentangle and modulate entangled semantics in the frequency domain.

\begin{figure*}[t]
    \centering
    \includegraphics[width=0.90\linewidth]{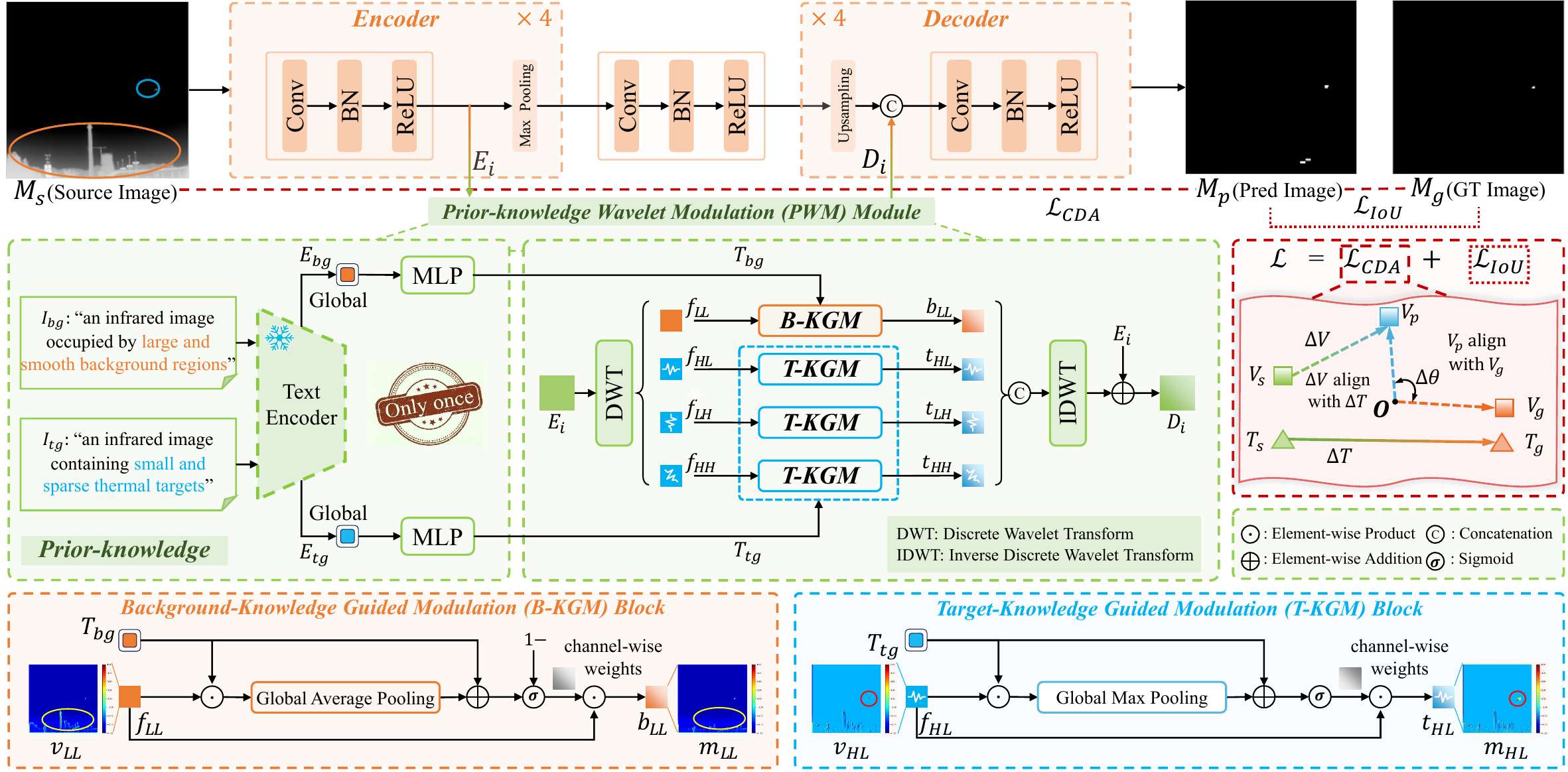}
    \vspace{-8pt}
    \caption{ Overview of our DGNet. DGNet adopts a four-stage encoder-decoder architecture, where each stage is equipped with a corresponding PWM module as the skip connection. In PWM module, the B-KGM block is designed to suppress background noise and the T-KGM block is designed to enhance target features, respectively. Finally, the network is jointly optimized using the Consensus-knowledge Directional Alignment (CDA) loss and the IoU loss.}
    \vspace{-12pt}
    \Description[DGNet architecture]{An overview of the DGNet framework with a four-stage encoder–decoder structure. Each stage includes a Prior-knowledge Wavelet Modulation (PWM) module composed of two branches: the Background Knowledge-Guided Modulation (BKGM) block, which suppresses background noise by modulating low-frequency components, and the Target Knowledge-Guided Modulation (TKGM) block, which enhances target features by refining high-frequency components. The modulated features are passed through the decoder to produce the final prediction, which is optimized using both the Consensus-knowledge Directional Alignment (CDA) loss and the IoU loss.}
    \label{fig:DGNet}
\end{figure*}

\subsection{Loss Functions for IRSTD}
In IRSTD, due to the extremely small target scale, detection performance is highly dependent on the employed loss function. Early methods, such as Binary Cross-Entropy (BCE) loss, supervise IRSTD in a pixel-wise manner by treating each pixel as an independent binary label. However, since targets occupy only a very small number of pixels, this loss fails to model target sparsity and often leads to missed targets.
To enhance the model’s focus on target regions, IoU~\cite{softiou} and Dice loss~\cite{diceloss} were introduced, which improve IRSTD performance by optimizing the overlap between predicted regions and ground truth. However, since large-scale targets contribute significantly more than small-scale ones, small targets tend to be neglected during optimization. Furthermore, SLS loss \cite{mshnet} enhances sensitivity to small targets by incorporating scale and location information, while FocalIoU \cite{mtunet} combines Focal loss~\cite{focal_loss} with IoU loss to suppress background responses and focus more on small targets. Nevertheless, these methods often suffer from instability and fail to generalize effectively across multi-scale target scenarios.

Essentially, these methods compute geometric discrepancies in the spatial domain, making them susceptible to gradient domination by large background regions. 
To address this, we design a CDA loss, which leverages cross-sample semantic consensus to construct a directed optimization path from the source state to the ideal state, significantly improving detection accuracy in complex scenarios.

\section{Method}
\label{sec:method}
\subsection{Overall Architecture}
The overall architecture of DGNet is illustrated in Fig. \ref{fig:DGNet}. DGNet is an end-to-end multimodal framework that adopts a four-stage encoder-decoder structure with skip connections. Each encoder stage extracts features and performs downsampling to obtain a larger receptive field, while the final encoder stage serves as a transition layer through convolutional blocks. The decoder progressively upsamples and refines feature maps to restore spatial resolution.
At these skip connections, we design a PWM module, which employs the Discrete Wavelet Transform (DWT) to decompose features into corresponding high/low-frequency subbands. Specifically, in the frequency domain, the PWM modulates the low-frequency components using a B-KGM block guided by background prior text, and modulates the high-frequency components using a T-KGM block guided by target prior text, thereby explicitly incorporating textual priors.
This process effectively suppresses large-scale smooth background clutter while enhancing discriminative target representations.
Finally, during training, the final prediction map $M_p$ is supervised by the CDA loss and IoU loss. The CDA loss introduces cross-sample consensus knowledge via textual guidance to construct a semantic optimization trajectory. Acting as a directional constraint in the feature space, it explicitly guides the network to evolve from a cluttered source state toward an ideal target-enhanced state, providing a unified evolution trajectory for model learning.

\subsection{PWM Module}
In complex IRSTD scenarios, severe background clutter makes it difficult for purely visual methods to effectively extract discriminative features between targets and backgrounds. The text-image fusion paradigm alleviates this issue by introducing semantic information. However, existing multi-modal methods typically rely on a single specific textual description. 
Due to the inherent structure of infrared images, which consists of large-scale smooth backgrounds and small-scale sparse targets, a single textual prompt often forces the network to process both background and target information simultaneously, leading to semantic entanglement that is difficult to disentangle. Moreover, reliance on image-specific textual prompts also limits practical deployment. 
To address this issue, we propose a Prior-knowledge Wavelet Modulation (PWM) module and embed it into the skip connections between the encoder and decoder. Specifically, in infrared images, targets usually exhibit small and sparse distributions, while background regions are typically large and smooth. The PWM module employs the DWT to decompose visual features into high/low-frequency subbands, and introduces fixed dual textual priors of targets and backgrounds in the frequency domain for explicit modulation, thereby enabling efficient separation of targets from complex background clutter.
\begin{figure*}[t]
    \centering
    \includegraphics[width=0.9\linewidth]{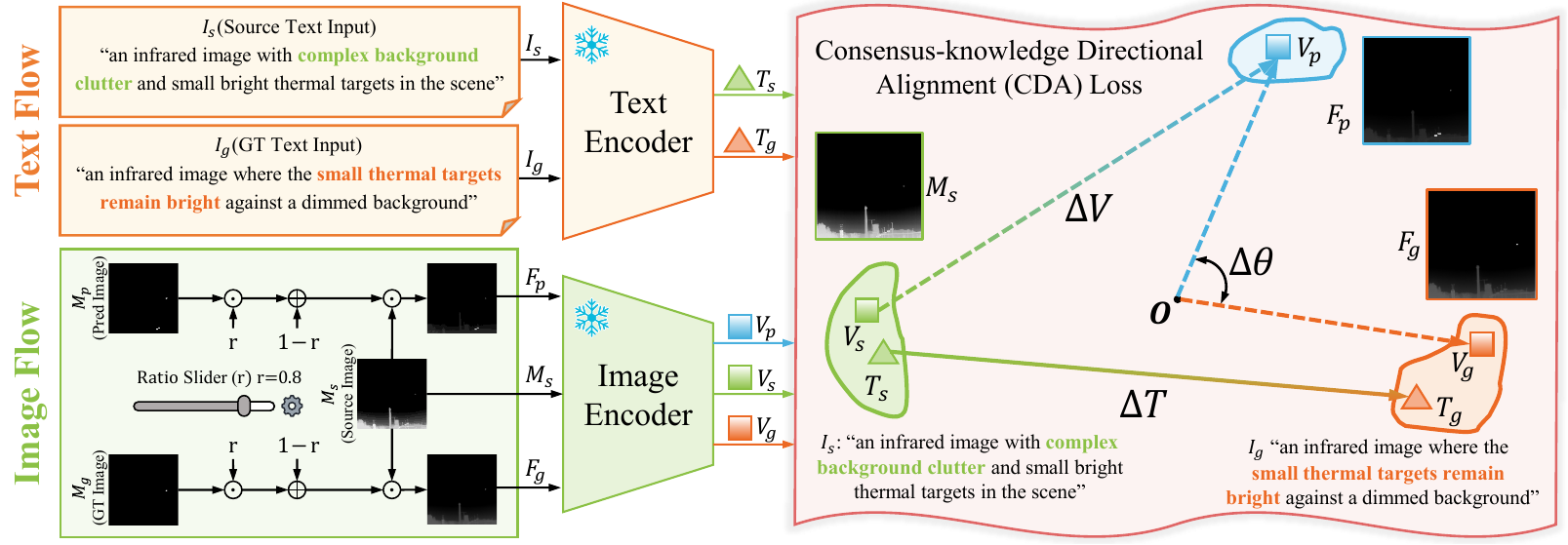}
    \vspace{-8pt}
    \caption{ Illustration of the Consensus-knowledge Directional Alignment (CDA) Loss. The CDA loss explicitly aligns the visual optimization path (\(\Delta V\)) with the semantic trajectory (\(\Delta T\)) derived from predefined consensus texts. By minimizing the angular deviation (\(\Delta \theta\)), it effectively forces the network to optimize from the cluttered source state (\(V_p\)) toward the ideal target state (\(V_g\)).}
    \vspace{-12pt}
    \Description[CDA loss alignment mechanism]{A schematic illustration of the Consensus-knowledge Directional Alignment (CDA) loss. The diagram shows a visual feature space where the current prediction state (\(V_p\)) and the ideal target state (\(V_g\)) define a visual optimization direction (\(\Delta V\)). In parallel, a semantic trajectory (\(\Delta T\)) is derived from predefined consensus texts. The CDA loss constrains the angular difference (\(\Delta \theta\)) between \(\Delta V\) and (\(\Delta T\)), encouraging the network to align its optimization path from a cluttered source state toward a target-enhanced representation.}
    \label{fig:CDA_Loss}
\end{figure*}

As illustrated in Fig. \ref{fig:DGNet}, the PWM module connects the corresponding encoder and decoder layers. Taking the output feature of the first encoder layer $E_1 \in \mathbb{R}^{256\times256\times16}$ as an example, it is first decomposed into one low-frequency subband and three high-frequency subbands via the DWT, as formulated in Eq.~\ref{eq:DWT}:
\begin{equation}
f_{LL},f_{HL},f_{LH},f_{HH} = DWT(E_1),
\label{eq:DWT}
\end{equation}
where $f_{LL}$ represents the low-frequency approximation component containing coarse structural information, while $f_{HL}$, $f_{LH}$, and $f_{HH}$ capture texture details along different orientations.

We introduce two fixed textual priors \(I_{bg}\) and \(I_{tg}\) based on the characteristics of infrared small-target images. These priors are mapped into global semantic embeddings \(E_{bg}, E_{tg} \in \mathbb{R}^D\) through a text encoder, and further projected into channel-wise modulation weights via a multi-layer perceptron (MLP), using Eq.~\ref{eq:MLP}:
\begin{equation}
	T_{bg} = MLP(\mathcal{F}(I_{bg})), \quad T_{tg} = MLP(\mathcal{F}(I_{tg})),
        \label{eq:MLP}
\end{equation}
where \(T_{bg}, T_{tg} \in \mathbb{R}^{C\times 1 \times 1}\). \(\mathcal{F}(\cdot)\) is the pre-trained language model CLIP with frozen weights, and since we adopt fixed textual prompts, only a single feature extraction is required. $T_{bg}$ inherently describes the low-frequency characteristics of the scene, while $T_{tg}$ corresponds to high-frequency characteristics, we use these semantic embeddings to modulate different frequency subband features along the channel dimension. Specifically, the low-frequency visual component is modulated with $T_{bg}$ via the B-KGM block to suppress large and smooth background, while the high-frequency visual components are modulated with $T_{tg}$ via the T-KGM block to emphasize sparse targets. Fig. \ref{fig:DGNet} illustrates the detailed structure of the B-KGM and T-KGM blocks. 
In B-KGM block, the visual feature $f_{LL}$ is modulated using the textual embedding $T_{bg}$ to obtain the background-suppressed feature $b_{LL}$:
\begin{equation}
	b_{LL} = (1-\sigma(GAP(T_{bg} \odot f_{LL}) + T_{bg})) \odot f_{LL},
        \label{eq:bkgm}
\end{equation}
where \(GAP(\cdot)\) denotes Global Average Pooling and $\sigma(\cdot)$ is the sigmoid function. In the B-KGM module shown in Fig.~\ref{fig:DGNet}, \(f_{LL}\) and \(b_{LL}\) are convolved and projected into a single channel, yielding the corresponding visual features \(v_{LL}\) and \(m_{LL}\). It can be observed that the background regions are effectively suppressed.
First, the $T_{bg}$ is projected into the same channel dimension as $f_{LL}$ and broadcast to match its spatial resolution. 
The term ($T_{bg} \odot f_{LL}$) models the global correlation between the background prior and the visual feature. 
Subsequently, the aggregated term \(GAP(\cdot) + T_{bg}\) captures the global background response strength. Through the \(1 - \sigma(\cdot)\) gating mechanism, channel-wise weights correlated with the background prior are adaptively suppressed, effectively reducing the influence of large-scale smooth background regions.
For the high-frequency visual feature, taking \(f_{HL}\) as an example, the corresponding target-enhanced feature $t_{HL}$ is obtained via:
\begin{equation}
	t_{HL} = \sigma(GMP(T_{tg}\odot f_{HL}) + T_{tg}) \odot f_{HL},
        \label{eq:tkgm}
\end{equation}
where \(GMP(\cdot)\) is the Global Max Pooling. In the T-KGM module shown in Fig.~\ref{fig:DGNet}, by applying convolution operations to \(f_{HL}\) and \(t_{HL}\), the targets in \(v_{HL}\) are effectively enhanced in \(m_{HL}\). $GMP(\cdot)$ captures globally salient responses, while $\sigma(\cdot)$ serves as an enhancement gating mechanism. Therefore, the target-related prior $T_{tg}$ highlights sparse and discriminative responses in high-frequency features, effectively enhancing target details. 
Similarly, we obtain the remaining high-frequency features $t_{LH}$ and $t_{HH}$ through T-KGM. These features are transformed back to the spatial domain via IDWT, and fused with the feature $E_1$ through a residual operation to obtain the prior text-modulated feature map $D_1$:
\begin{equation}
	D_1 = E_1 + IDWT(cat(b_{LL},t_{HL},t_{LH},t_{HH})).
        \label{eq:IDWT}
\end{equation}
By leveraging prior knowledge of targets and backgrounds, the PWM module explicitly guides the network to decouple targets from complex background clutter, providing highly discriminative features for subsequent decoding stages.
\begin{table*}[!t]
	\renewcommand{\arraystretch}{0.9}
        \setlength{\tabcolsep}{9.88pt}
        \caption{Quantitative comparisons between our DGNet and 21 state-of-the-art (SOTA) methods on the IRSTD-1K, SIRST, NUDT-SIRST datasets in terms of IoU(\%), ${\rm P_d}$(\%) and ${\rm F_a}$($10^{-6}$). The best results are in bold. In the Type column, `Trad' denotes traditional methods, `Purely-V' denotes purely visual methods, and `Text-V' denotes text-image fusion methods.} 
        \vspace{-8pt}
        \resizebox{0.97\linewidth}{!}{
		\begin{tabular}{lcccccccccc}
        \Xhline{1.5pt}
		\multicolumn{1}{c}{\multirow{2}{*}{Method}} & \multicolumn{3}{c}{{IRSTD-1K}} & \multicolumn{3}{c}{{SIRST}} & \multicolumn{3}{c}{NUDT-SIRST} & \multicolumn{1}{c}{\multirow{2}{*}{Type}}\\
		\cmidrule(lr){2-4} \cmidrule(lr){5-7} \cmidrule(lr){8-10}
		& IoU$\uparrow$ & ${\rm P_d}\uparrow$ & ${\rm F_a}\downarrow$ & IoU$\uparrow$ & ${\rm P_d}\uparrow$ & ${\rm F_a}\downarrow$& IoU$\uparrow$ & ${\rm P_d}\uparrow$ & ${\rm F_a}\downarrow$ & \\
        \hline 
        PSTNN \cite{low_rank_4} (RS'19) & 24.57 & 71.99 & 35.26 & 30.30 & 72.80 & 48.99 & 14.85 & 66.13 & 44.17 & \multirow{4}{*}{Trad}\\ 
        TLLCM \cite{contrast_1} {(GRSL'19)} & 3.31 & 77.39 & 6738 & 4.24 & 88.37 & 6243 & 2.18 & 62.01 & 1608 & \\ 
        MSLSTIPT \cite{low_rank_3} {(TGRS'20)} & 11.43 & 79.03 & 1524 & 1.08 & 0.05 & 8.18 & 8.34 & 47.40 & 888.1 & \\
        WSLCM \cite{contrast_2} {(GRSL'20)} & 3.45 & 72.44 & 6619 & 6.39 & 88.74 & 4462 & 2.28 & 56.82 & 1309 & \\
        \hline
        MDvsFA \cite{mdvsfa} {(ICCV'19)} & 37.34 & 83.71 & 88.52 & 60.30 & 89.35 & 56.35 & 35.86 & 85.22 & 95.37 & \multirow{16}{*}{Purely-V}\\
		  ALCNet \cite{alcnet} {(TGRS'21)} & 65.68 & 89.25 & 27.71 & 73.74 & 97.25 & 26.79 & 72.89 & 96.19 & 30.40 & \\
		  ACMNet \cite{acmnet} {(WACV'21)} & 60.33 & 93.27 & 68.49 & 69.44 & 92.02 & 22.71 & 64.86 & 96.72 & 28.59 & \\
		  ISNet \cite{isnet} {(CVPR'22)} & 61.85 & 90.24 & 31.56 & 70.49 & 95.06 & 67.98 & 81.24 & 97.78 & 6.34 & \\
        DNANet \cite{dnanet} {(TIP'22)} & 65.71 & 91.84 & 17.61 & 77.76 & 96.33 & 10.29 & 88.19 & 98.62 & 9.00 & \\
		  UIU-Net \cite{uiunet} {(TIP'23)} & {68.69} & 91.25 & 13.48 & 77.53 & 92.40 & 9.33 & 75.91 & 96.83 & 18.61 & \\
		RPCANet \cite{rpcanet} {(WACV'23)} & 63.21 & 88.31 & {4.39} & 65.08 & 93.58 & 10.85 & {89.31} & 97.14 & 2.87 & \\
		SCTransNet \cite{sctransnet} {(TGRS'24)} & 68.03 & 93.27 & {10.74} & 77.50 & 96.95 & 13.92 & {94.09} & {98.62} & 4.29 & \\
        PBT \cite{pbt} {(TGRS'24)} & 68.49 & 92.52 & 8.88 & 78.39 & 99.08 & 2.13 & 83.89 & 97.23 & 4.23 & \\
        MSHNet \cite{mshnet} {(CVPR'24)} & 67.68 & 92.89 & 12.69 & 73.50 & 97.25 & 31.05 & 80.55 & 97.99 & 11.77 & \\
        GSFANet \cite{gsfanet} {(TGRS'25)} & 68.60 & 91.84 & 11.01 & 73.58 & 98.17 & 11.71 & 93.96 & {99.05} & 4.07 & \\
        BGM \cite{bgm} {(TGRS'25)} & 69.23 & 91.50 & 11.39 & 76.17 & 98.17 & 12.42 & 93.33 & 98.84 & 5.86 & \\
        DRPCA-Net \cite{drpcanet} {(TGRS'25)} & 66.33 & 91.07 & 16.93 & 72.82 & 98.77 & 9.23 & {93.33} & 99.15 & 6.05 & \\
        IRPNet \cite{irpnet} {(TGRS'26)} & 68.97 & 91.84 & 7.52 & 79.19 & 99.08 & 6.74 & {93.65} & 98.31 & 3.65 & \\
        PQGNet \cite{pqgnet} {(TGRS'26)} & 69.88 & 92.78 & 6.68 & 80.61 & 99.08 & 13.72 & {93.67} & 98.41 & 7.35 & \\
        FGARNet \cite{fgarnet} {(TGRS'26)} & 70.30 & 91.16 & 14.42 & 78.47 & 98.15 & 3.37 & 93.52 & 98.72 & 1.97 & \\
        \hline
        SAIST \cite{saist} {(CVPR'25)} & {72.14} & \textbf{96.18} & 4.76 & {80.82} & {99.56} & \textbf{0.87} & {95.23} & {99.28} & {1.31} & \multirow{2}{*}{Text-V}\\
        
		\cellcolor[HTML]{E0E9F7}{\textbf{DGNet(Ours)}} & \cellcolor[HTML]{E0E9F7}\textbf{72.72} & \cellcolor[HTML]{E0E9F7}{93.88} & \cellcolor[HTML]{E0E9F7}\textbf{4.25} & \cellcolor[HTML]{E0E9F7}\textbf{82.68} & \cellcolor[HTML]{E0E9F7}\textbf{100} & \cellcolor[HTML]{E0E9F7}{1.24} & \cellcolor[HTML]{E0E9F7}\textbf{95.78} & \cellcolor[HTML]{E0E9F7}\textbf{99.37} & \cellcolor[HTML]{E0E9F7}\textbf{1.19} & \\
        \Xhline{1.5pt}
		\end{tabular}
        }
    \vspace{-12pt}
	\label{table:main_quantitative_result}
\end{table*} 

\subsection{CDA Loss}
\subsubsection{Motivation and Semantic Consensus Formulation}
In IRSTD, existing pixel-level losses, such as IoU losses, primarily enforce local geometric consistency in the spatial domain. However, they lack high-level semantic guidance, making the model sensitive to complex background noise and leading to unstable optimization and limited generalization. Meanwhile, existing test-image methods construct image-specific textual descriptions for each image, which introduces additional inference overhead.

To address this issue, we explore the consensus across the optimization processes of different images and attempt to model the optimization objective using natural language~\cite{clip_loss}. Specifically, we model the initial state of all samples as `complex background'  and the ideal optimization endpoint as `bright targets', forming cross-sample consensus knowledge.
Based on this insight, we unify the initial state of all samples as a consensus source text prompt \(I_s\): `an infrared image with complex background clutter and small thermal targets'. Correspondingly, the ideal optimization endpoint is defined as a consensus target text prompt \(I_g\): `an infrared image where small thermal targets remain bright against a dimmed background'.

\subsubsection{Semantic and Visual Trajectories in CLIP Space}
To mathematically characterize the above semantic transition while avoiding complex analytical modeling, we introduce the pre-trained vision-language model CLIP, which aligns visual and textual modalities in a shared embedding space.

First, as illustrated in Fig.~\ref{fig:CDA_Loss}, we feed the fixed consensus texts \(I_{s}\) and \(I_{g}\) into the frozen CLIP text encoder to obtain their corresponding semantic embeddings: \(T_s, T_g\). Based on this, the semantic optimization direction in the embedding space is defined as:
\begin{equation}
\Delta T = T_g - T_s,
\end{equation}
this vector characterizes the semantic transition from `background clutter' to `salient small targets', thereby constructing a \textbf{global semantic optimization trajectory} that is independent of specific image content. Meanwhile, we need to map visual states into the CLIP embedding space. However, since CLIP struggles to encode prediction maps or ground-truth masks that lack structural information, we propose a mask-guided image fusion strategy. This strategy aims to provide CLIP with structurally informative visual inputs, compensating for its limitations in encoding sparse masks and enabling effective alignment between the predicted and ideal states in the semantic space. 
Specifically, we fuse the predicted map \(M_p\) and the ground-truth map \(M_g\) with the original infrared image \(M_s\) to generate the CLIP-encodable predicted visual state \(F_p\) and the ideal visual state \(F_g\), respectively, formulated as follows,
\begin{equation}
F_p = (r \odot M_p + 1 - r) \odot M_s, F_g = (r \odot M_g + 1 - r) \odot M_s,
\end{equation}
where ratio slider \(r = 0.8\) is a hyperparameter controlling the degree of background suppression, and \(\odot\) denotes element-wise multiplication. Subsequently, the \(F_p\), \(M_s\), and \(F_g \) are fed into the frozen CLIP image encoder to extract visual embeddings \(V_p\), \(V_s\), and \(V_g\), respectively.
During the training phase, we define the transition from the source feature \(V_s\) to the predicted feature \(V_p\) as the visual feature evolution direction, formulated as:
\begin{equation}
\Delta V = V_{p} - V_{s},
\end{equation}
\subsubsection{Consensus-knowledge Directional Alignment Loss}
To ensure that the model's optimization follows the expected semantic objective, we constrain the visual change direction \(\Delta V\) to be consistent with the semantic direction \(\Delta T\). Specifically, we enforce the visual feature evolution to be parallel to the semantic optimization direction. As illustrated in Fig.~\ref{fig:CDA_Loss}, the dashed arrow from \(V_s\) to \(V_p\) is constrained to align with the solid arrow from \(T_s\) to \(T_g\). This alignment guides the detection process toward the language-defined trajectory, reaching the desired state in the embedding space. Therefore, we define the Consensus-knowledge Directional (CD) loss as:
\begin{equation}
\mathcal{L}_{CD} = \frac{1}{2}(1 - \frac{\Delta V \cdot \Delta T}{\|\Delta V\| \cdot \|\Delta {T}\|}),
\end{equation}
where \(\|\cdot\|\) denotes the \(L2\) norm, which measures the length of a vector. Additionally, to directly constrain the alignment between the predicted visual feature and the ideal visual state, we define the Consensus-knowledge Alignment (CA) loss:
\begin{equation}
\mathcal{L}_{CA} = \frac{1}{2}(1 - cos(\angle \theta))=\frac{1}{2}(1 - \frac{ V_p \cdot V_g}{\|V_p\| \cdot \|V_g\|})
\end{equation}

Combining both constraints, the final \(\mathcal{L}_{CDA}\) is defined as:
\begin{equation}
\mathcal{L}_{CDA} = \frac{1}{2}\mathcal{L}_{CD} + \frac{1}{2}\mathcal{L}_{CA}.
\end{equation}

The design of \(\mathcal{L}_{CDA}\) provides a high-level semantic optimization strategy for IRSTD. Finally, to further enhance model performance, the ultimate training loss is defined as \(\mathcal{L}\):
\begin{equation}
\mathcal{L} = \mathcal{L}_{CDA} + \mathcal{L}_{IoU},
\end{equation}
this design ensures robust pixel-level learning while guiding the model toward the desired high-level semantic state.
\begin{figure*}[t]
    \centering
    \includegraphics[width=0.87\linewidth]{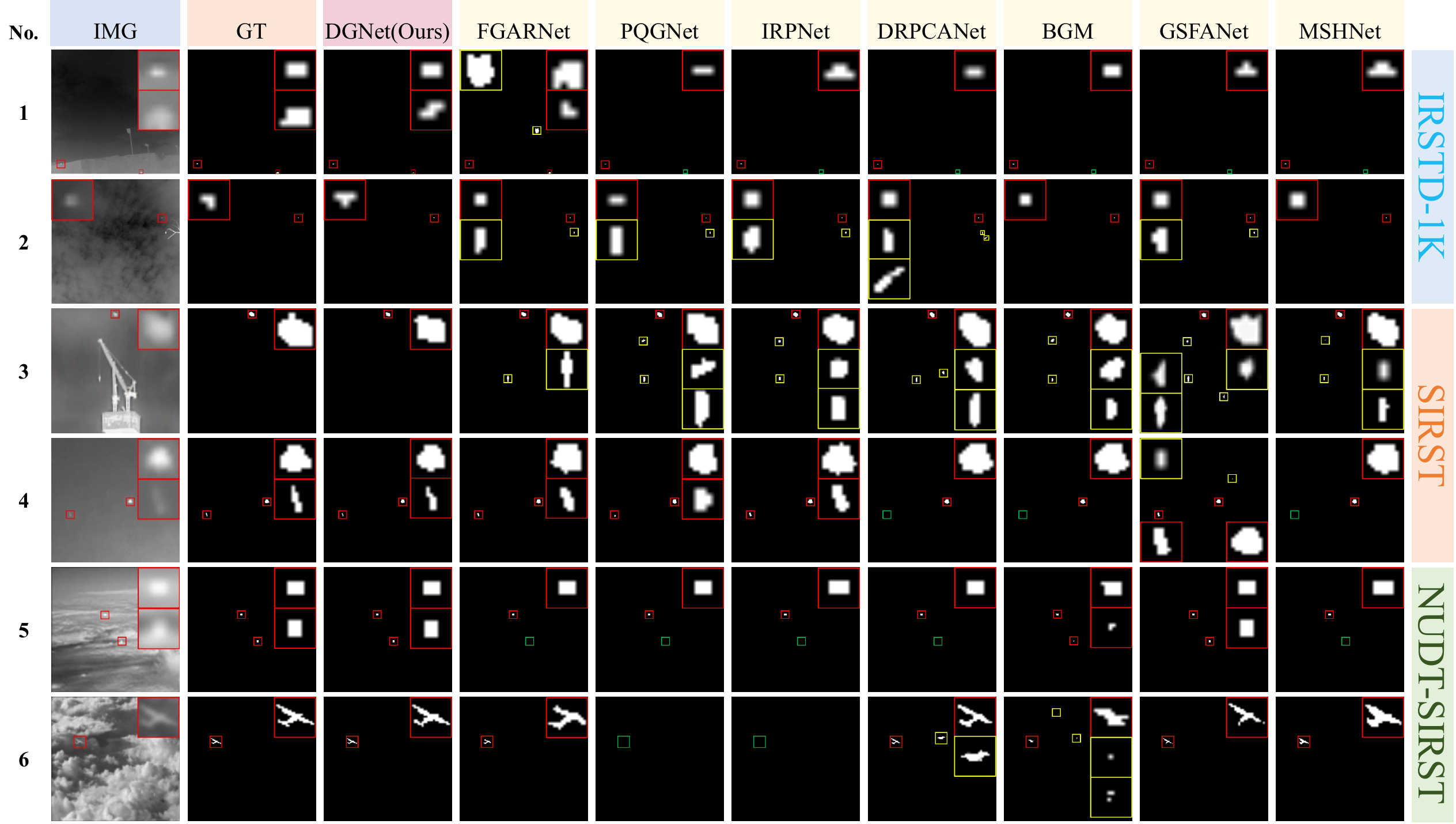}
    \vspace{-8pt}
    \caption{ Visual results of different IRSTD methods. The boxes in red, yellow, and green represent correct, false alarms and missed targets, respectively. The enlarged views are shown in the corners.}
    \vspace{-13pt}
    \Description[Qualitative comparison]{Qualitative comparison of infrared small target detection results across multiple methods on different scenes. Each row corresponds to a sample infrared image, while each column shows the detection results of a specific method. Red boxes indicate correctly detected targets, yellow boxes denote false alarms, and green boxes represent missed targets. Zoomed-in regions are provided in the corners to highlight fine-grained differences in target localization and background suppression.}
    \label{fig:sota_exp}
\end{figure*}

\begin{figure}[t]
    \centering
    \includegraphics[width=0.81\linewidth]{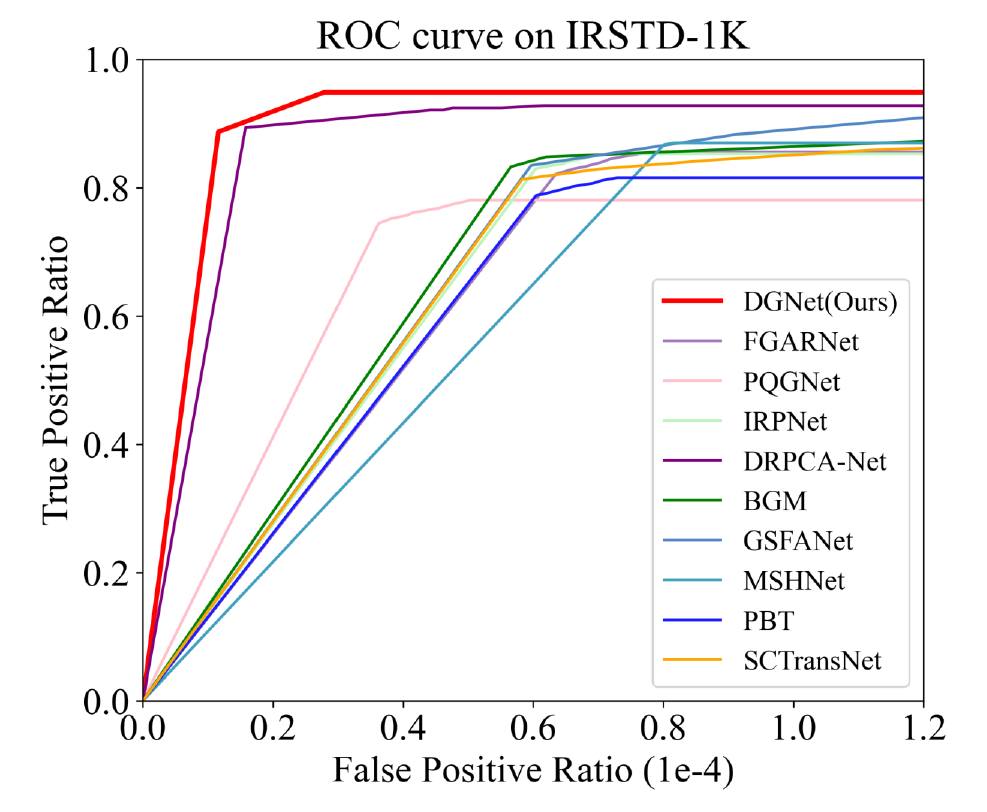}
    \vspace{-8pt}
    \caption{ROC curve on the IRSTD-1K dataset. }
    \vspace{-15pt}
    \Description[ROC curves of different methods on the IRSTD-1K dataset.]{ROC curves of different methods on the IRSTD-1k. The closer curves to the top-left corner, the better performance.}
    \label{fig:roc}
\end{figure}

\section{Experiments}
\label{sec:experiments}
\subsection{Datasets and Evaluation Metrics}
\textbf{Datasets:} All experiments are conducted on three widely used datasets: IRSTD-1K~\cite{isnet}, SIRST~\cite{acmnet}, and NUDT-SIRST~\cite{dnanet}, which contain 1001, 427, and 1327 infrared images, respectively. Following existing works~\cite{isnet,hdnet}, the images in IRSTD-1K and SIRST are split into training and testing sets with a 4:1 ratio, while NUDT-SIRST is divided with 50\% for training and 50\% for testing.

\textbf{Evaluation Metrics:} We adopt several widely used metrics to evaluate our proposed DGNet and existing methods, including Intersection over Union (\(IoU\)) for pixel-level evaluation, as well as Probability of Detection (\(P_d\)) and False Alarm Rate (\(F_a\)) for object-level evaluation. In addition, we plot Receiver Operating Characteristic (ROC) curves based on different True Positive Rates (TPR) and False Positive Rates (FPR).

\subsection{Implementation Details}
Our DGNet is implemented with the PyTorch framework on a single NVIDIA GeForce RTX 4090 GPU. The model is trained for 600 epochs with a batch size of 16, utilizing the Adam optimizer. 
The initial learning rate is set to 5e-4 and decayed by a factor of 0.9 at epochs 300 and 450.
The input images are resized to \(256 \times 256\).
During training, we adopt the CLIP-ViT-B/32~\cite{clip} as the text and image encoders, while it is not used during inference, incurring no additional overhead. For comparison, we evaluate our DGNet with 21 SOTA methods on three challenging datasets. 
For fairness, all quantitative and qualitative results are either taken from the authors’ public results or reproduced using their released code.

\subsection{Quantitative Comparison}
Table \ref{table:main_quantitative_result} presents the quantitative comparison of different methods on three public datasets, including IRSTD-1K, SIRST, and NUDT-SIRST. Our DGNet demonstrates significant performance improvements. 
Specifically, on the most challenging IRSTD-1K dataset, our method achieves the highest \(IoU\) of 72.72\%, significantly outperforming existing approaches, while reducing the \(F_a\) to 4.25. Meanwhile, DGNet maintains a high \(Pd\), demonstrating its strong capability to effectively extract small targets in complex backgrounds. 
On the SIRST dataset, DGNet achieves a \(Pd\) of 100\%, along with an \(IoU\) of 82.68\% and a low \(F_a\) of 1.24, which verifies that our method can achieve accurate and complete target segmentation in complex scenarios. 
Furthermore, by modulating visual features with prior knowledge and consensus knowledge, DGNet effectively suppresses background clutter and extracts complete small targets. On the NUDT-SIRST dataset, DGNet achieves an \(IoU\) as high as 95.78\% and a \(Pd\) of 99.37\%. 
Although our method is slightly inferior to SAIST~\cite{saist} in terms of \(Pd\) on IRSTD-1K and \(F_a\) on SIRST, DGNet does not require complex image-specific text design. Instead, by leveraging dual knowledge to modulate visual features, our model exhibits stronger robustness and generalization ability.

In addition, we present the ROC curves of different IRSTD methods on the IRSTD-1K dataset in Fig.~\ref{fig:roc}. The results show that DGNet achieves a higher TPR at lower FPR, demonstrating its strong competitiveness compared to other SOTA methods.

\subsection{Qualitative Comparison}
Fig.~\ref{fig:sota_exp} presents qualitative comparisons between DGNet and seven representative methods under various challenging scenarios. It can be observed that purely vision-based detection methods (e.g., FGARNet) are affected by complex backgrounds, resulting in false alarms (rows 1-3, column 4). Under conditions such as dense cloud occlusion and extremely low signal-to-noise ratios, most detection methods struggle to extract discriminative features between targets and background, leading to missed targets.
In contrast, DGNet benefits from precise modulation of the PWM module in the frequency domain. By leveraging two textual priors that describe the background as `large and smooth' and the targets as `small and sparse', the model explicitly suppresses background clutter while enhancing target responses. Combined with the cross-sample consistent optimization direction provided by the CDA loss, DGNet effectively separate targets from the background, demonstrating strong generalization ability and robustness.

\begin{table}[t]
    \centering
    \renewcommand{\arraystretch}{0.90}
    \setlength{\tabcolsep}{4.68pt}
    \caption{Ablation study of PWM module and CDA loss.}
    \vspace{-8pt}
	\begin{center}
        {\fontsize{6pt}{6pt}}\selectfont
		\begin{tabular}{lcccccc}
			\hline
			\multirow{2}{*}{Variants} & \multicolumn{3}{c}{{IRSTD-1k}} & \multicolumn{3}{c}{{SIRST}}\\
			\cmidrule(lr){2-4} \cmidrule(lr){5-7}
			& IoU$\uparrow$ & ${\rm P_d}\uparrow$ & ${\rm F_a}\downarrow$& IoU$\uparrow$ & ${\rm P_d}\uparrow$ & ${\rm F_a}\downarrow$\\
			\hline
		    base & 63.17 & 88.46 & 20.88 & 74.07 & 95.41 & 26.08 \\
            base+PWM & 69.87 & 91.16 & 16.17 & 78.48 & 97.25 & 10.47 \\
            base + \(\mathcal{L}_{CDA}\) & 69.90 & 92.52 & 11.77 & 79.54 & 98.17 & 8.34 \\
            \hline
            \cellcolor[HTML]{E0E9F7}{\textbf{DGNet (Ours)}}  & \cellcolor[HTML]{E0E9F7}{\textbf{72.72}} & \cellcolor[HTML]{E0E9F7}{\textbf{93.88}} & \cellcolor[HTML]{E0E9F7}{\textbf{4.25}} & \cellcolor[HTML]{E0E9F7}{\textbf{82.68}} & \cellcolor[HTML]{E0E9F7}{\textbf{100}} & \cellcolor[HTML]{E0E9F7}{\textbf{1.24}} \\
			\hline
		\end{tabular}
	\label{ablation_module}
	\vspace{-12pt}
	\end{center}
\end{table}

\begin{figure}[t]
    \centering
    \includegraphics[width=0.9\linewidth]{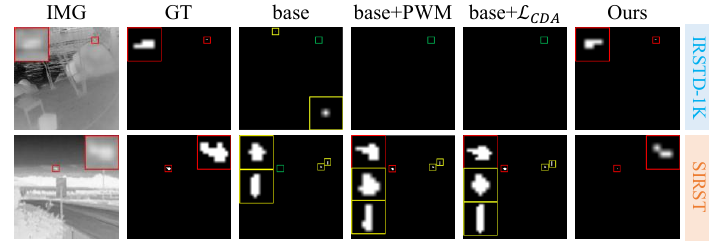}
    \vspace{-8pt}
    \caption{ Visual examples of ablation experiments between PWM module and CDA loss.}
    \vspace{-10pt}
    \Description[Visual examples of ablation experiments between PWM module and CDA loss.]{Qualitative ablation results comparing the effects of the PWM module and the CDA loss on infrared small target detection.}
    \label{fig:ablation_intra_exp}
\end{figure}

\subsection{Ablation Study}
\subsubsection{Ablation Experiments Between PWM module and CDA loss}
To verify the effectiveness of the PWM module and CDA loss in DGNet, we conduct comprehensive ablation studies on the IRSTD-1K and SIRST datasets. A standard encoder-decoder architecture is adopted as the baseline model (base), upon which the PWM module and CDA loss are progressively introduced.
As shown in Table~\ref{ablation_module}, without dual-knowledge guidance, the purely visual baseline exhibits extremely high $F_a$ on both datasets and achieves the worst performance in terms of $IoU$ and $P_d$.
When the PWM module is incorporated into the `base', all performance metrics are significantly improved. 
Likewise, introducing the CDA loss brings substantial performance gains, particularly in suppressing \(F_a\). 
When both the PWM module and CDA loss are integrated, the full DGNet achieves the best overall performance and significantly outperforms the `base'. 
Furthermore, as shown in Fig.~\ref{fig:ablation_intra_exp}, the `base' model exhibits severe false alarms and missed targets. With the equipment of both the PWM module and CDA loss, the full DGNet effectively suppresses background interference, while accurately detecting small targets.
These results clearly demonstrate that the PWM module effectively modulates target features against complex backgrounds, while the CDA loss constructs a cross-sample semantic optimization trajectory in the CLIP embedding space, providing a clear and reliable learning direction for the model.

\subsubsection{Impact of the PWM module}
To analyze the contributions of the key components within the PWM module, we conducted detailed internal ablation studies on the IRSTD-1K and SIRST datasets. As shown in Table \ref{ablation_PWM}, when the DWT is removed and feature modulation is performed only in the spatial domain (`w/o wave’), the model performance drops noticeably. 
This is because the DWT separates high-frequency edges from low-frequency smooth components, providing a decoupled representation space for subsequent text-guided modulation.
Removing the T-KGM block weakens the model’s ability to detect faint targets. 
On the other hand, the B-KGM block plays a critical role in controlling false alarms. 
In addition, Fig.~\ref{fig:ablation_inter_exp}(a) shows the visual results of several model variants. It is noteworthy that removing the T-KGM block leads to obvious missed targets (column 4). Similarly, removing the B-KGM block results in false alarms (row 1, column 5). Equipped with the full PWM module, DGNet effectively detects targets while suppressing background interference.
In summary, the PWM module constructs a frequency-decoupled representation space via the wavelet transform, where the T-KGM branch enhances high-frequency target features and the B-KGM branch suppresses low-frequency background clutter.

\begin{table}[t]
    \centering
    \renewcommand{\arraystretch}{0.9}
    \setlength{\tabcolsep}{4.68pt}
    \caption{Ablation study of PWM module.}
    \vspace{-8pt}
	\begin{center}
        {\fontsize{6pt}{6pt}}\selectfont
		\begin{tabular}{lcccccc}
			\hline
			\multirow{2}{*}{Variants} & \multicolumn{3}{c}{{IRSTD-1k}} & \multicolumn{3}{c}{{SIRST}}\\
			\cmidrule(lr){2-4} \cmidrule(lr){5-7}
			& IoU$\uparrow$ & ${\rm P_d}\uparrow$ & ${\rm F_a}\downarrow$& IoU$\uparrow$ & ${\rm P_d}\uparrow$ & ${\rm F_a}\downarrow$\\
			\hline
		    w/o wave & 70.12 & 92.52 & 11.24 & 80.38 & 98.17 & 6.03 \\
            w/o T-KGM & 71.04 & 91.50 & 7.74 & 81.08 & 98.17 & 7.45 \\
            w/o B-KGM & 71.47 & 93.20 & 10.70 & 81.59 & 99.08 & 12.95 \\
            \hline
            \cellcolor[HTML]{E0E9F7}{\textbf{DGNet (Ours)}} & \cellcolor[HTML]{E0E9F7}{\textbf{72.72}} & \cellcolor[HTML]{E0E9F7}{\textbf{93.88}} & \cellcolor[HTML]{E0E9F7}{\textbf{4.25}} & \cellcolor[HTML]{E0E9F7}{\textbf{82.68}} & \cellcolor[HTML]{E0E9F7}{\textbf{100}} & \cellcolor[HTML]{E0E9F7}{\textbf{1.24}} \\
			\hline
		\end{tabular}
	\label{ablation_PWM}
	\vspace{-12pt}
	\end{center}
\end{table}

\begin{table}[t]
    \centering
    \renewcommand{\arraystretch}{0.9}
    \setlength{\tabcolsep}{4.68pt}
    \caption{Ablation study of CDA Loss.}
    \vspace{-9pt}
	\begin{center}
        {\fontsize{6pt}{6pt}}\selectfont
		\begin{tabular}{lcccccc}
			\hline
			\multirow{2}{*}{Variants} & \multicolumn{3}{c}{{IRSTD-1k}} & \multicolumn{3}{c}{{SIRST}}\\
			\cmidrule(lr){2-4} \cmidrule(lr){5-7}
			& IoU$\uparrow$ & ${\rm P_d}\uparrow$ & ${\rm F_a}\downarrow$& IoU$\uparrow$ & ${\rm P_d}\uparrow$ & ${\rm F_a}\downarrow$\\
			\hline
		    base\(_p\) & 69.87 & 91.16 & 16.17 & 78.48 & 97.25 & 10.47 \\
            base\(_p\) + \(\mathcal{L}_{CD}\) & 71.86 & 92.18 & 10.25 & 81.46 & 98.17 & 8.16 \\
            base\(_p\) + \(\mathcal{L}_{CA}\) & 71.37 & 92.52 & 9.64 & 81.12 & 99.08 & 7.63 \\
            \hline
            \cellcolor[HTML]{E0E9F7}{\textbf{DGNet (Ours)}} & \cellcolor[HTML]{E0E9F7}{\textbf{72.72}} & \cellcolor[HTML]{E0E9F7}{\textbf{93.88}} & \cellcolor[HTML]{E0E9F7}{\textbf{4.25}} & \cellcolor[HTML]{E0E9F7}{\textbf{82.68}} & \cellcolor[HTML]{E0E9F7}{\textbf{100}} & \cellcolor[HTML]{E0E9F7}{\textbf{1.24}} \\
			\hline
		\end{tabular}
	\label{ablation_CDA}
	\vspace{-12pt}
	\end{center}
\end{table}

\subsubsection{Impact of the CDA Loss}
\begin{figure}[t]
    \centering
    \includegraphics[width=0.86\linewidth]{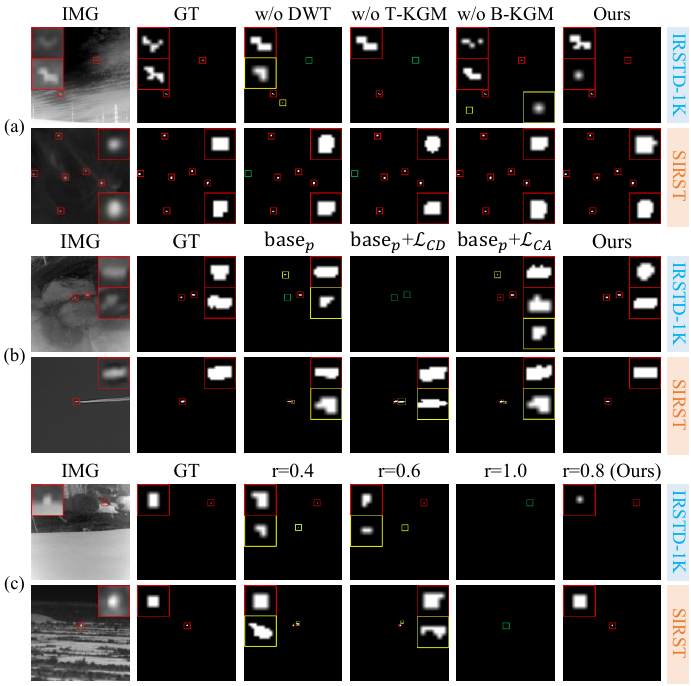}
    \vspace{-8pt}
    \caption{ Visual examples of ablation experiments inside PWM module, CDA loss and ratio slider (r) in CDA loss.}
    \vspace{-10pt}
    \Description[Visual examples of ablation experiments inside PWM module and CDA loss.]{Qualitative ablation results illustrating the internal effectiveness of different components within the PWM module, the CDA loss and the Ratio Slider (r) in \(\mathcal{L}_{CDA}\).}
    \label{fig:ablation_inter_exp}
\end{figure}

To further investigate the roles of each constraint term in the \(\mathcal{L}_{CDA}\), we conduct detailed ablation studies on the IRSTD-1K and SIRST datasets, and the quantitative results are shown in Table \ref{ablation_CDA}. We adopt the network with the PWM module as the baseline (`base\(_p\)') and use \(\mathcal{L}_{IoU}\) as the optimization function. When \(\mathcal{L}_{CD}\) is introduced, this loss provides a semantic optimization trajectory for the model, and the directional constraint it offers leads to steady performance improvements. Similarly, equipping the model with \(\mathcal{L}_{CD}\) aligns the predicted visual features with the ground-truth features in the CLIP space, enabling precise target detection. 
When the model is equipped with the full \(\mathcal{L}_{CDA}\) loss, DGNet achieves the best detection performance on both  datasets. Fig.~\ref{fig:ablation_inter_exp}(b) shows the visualization results of different optimization strategies, where the `base\(_p\)' obtains the poorest detection results. With the inclusion of \(\mathcal{L}_{CDA}\), the model significantly reduces false alarms and missed targets, fully recognizing small targets in complex images. This further demonstrates that \(\mathcal{L}_{CDA}\) provides the model with a high-level semantic optimization path, leveraging cross-sample semantic consensus, significantly improving detection accuracy in complex scenarios.

\begin{table}[t]
    \centering
    \renewcommand{\arraystretch}{0.88}
    \setlength{\tabcolsep}{5.18pt}
    \caption{Ablation study of Ratio Slider.}
    \vspace{-8pt}
	\begin{center}
        {\fontsize{6pt}{6pt}}\selectfont
		\begin{tabular}{lcccccc}
			\hline
			\multirow{2}{*}{Variants} & \multicolumn{3}{c}{{IRSTD-1k}} & \multicolumn{3}{c}{{SIRST}}\\
			\cmidrule(lr){2-4} \cmidrule(lr){5-7}
			& IoU$\uparrow$ & ${\rm P_d}\uparrow$ & ${\rm F_a}\downarrow$& IoU$\uparrow$ & ${\rm P_d}\uparrow$ & ${\rm F_a}\downarrow$\\
			\hline
		    r=0.4 & 70.75 & 90.14 & 13.82 & 80.38 & 97.25 & 8.88 \\
            r=0.6 & 71.66 & 92.86 & 9.72 & 82.09 & 99.08 & 5.68 \\
            r=1.0 & 71.16 & 93.54 & 8.96 & 81.64 & 98.16 & 4.97 \\
            \hline
            \cellcolor[HTML]{E0E9F7}{\textbf{r=0.8 (Ours)}} & \cellcolor[HTML]{E0E9F7}{\textbf{72.72}} & \cellcolor[HTML]{E0E9F7}{\textbf{93.88}} & \cellcolor[HTML]{E0E9F7}{\textbf{4.25}} & \cellcolor[HTML]{E0E9F7}{\textbf{82.68}} & \cellcolor[HTML]{E0E9F7}{\textbf{100}} & \cellcolor[HTML]{E0E9F7}{\textbf{1.24}} \\
			\hline
		\end{tabular}
	\label{ablation_Ratio}
	\vspace{-12pt}
	\end{center}
\end{table}

\subsubsection{Impact of the Ratio Slider (r) in \(\mathcal{L}_{CDA}\)}
In the CDA loss, the ratio slider $r$ determines the fusion result between the predicted map, the ground-truth map, and the original image, directly affecting the prominence of the background in the optimization target. 
To analyze the impact of $r$ on CLIP image encoding, we conduct comparative experiments with $r \in {0.4, 0.6, 0.8, 1.0}$, and the quantitative results are summarized in Table~\ref{ablation_Ratio}.
When $r$ decreases from 0.8 to 0.4, the background in the optimization target remains overly prominent. The encoded features are more susceptible to background clutter, leading to false alarms. 
When $r$=$1.0$, the optimization target completely loses background structure, and the image degrades into an entirely dark scene with only sparse bright spots. This severely impairs the ability of CLIP to encode image features, resulting in suboptimal model performance.
Experiments show that at $r$=$0.8$, the background is sufficiently darkened to highlight small targets while retaining weak global structural information. 
Under this configuration, DGNet achieves the best results on both datasets.
Similarly, Fig.~\ref{fig:ablation_inter_exp}(c) shows consistent detection results, when $r$=$0.4$ and $r$=$0.6$, the predicted images exhibit obvious false alarms, whereas at $r$=$1.0$, noticeable missed targets occur.
Overall, selecting an appropriate value for the ratio slider $r$ provides the image encoder with necessary contextual cues, maintaining the stability of the feature space.

\begin{table}[t]
    \centering
    \renewcommand{\arraystretch}{0.88}
    \setlength{\tabcolsep}{2.98pt}
    \caption{
    Comparison of model complexity between our DGNet and SOTA methods from the past three years.
    }
    \vspace{-8pt}
	\begin{center}
        {\fontsize{6pt}{6pt}}\selectfont
		\begin{tabular}{ccccc}
			\hline
			Method & Year & Params(M) {$\downarrow$} & FLOPs(G) {$\downarrow$} & FPS(f/s) {$\uparrow$} \\
			\hline
                SCTransNet~\cite{sctransnet} & 2024 & 11.19 & 20.24 & 36.19 \\
                PBT~\cite{pbt} & 2024 & 26.29 & 28.53 & 16.67 \\
                MSHNet~\cite{mshnet} & 2024 & 4.07 & 6.11 & 80.12 \\
                GSFANet~\cite{gsfanet} & 2025 & 2.97 & 5.25 & 25.87 \\
                BGM~\cite{bgm} & 2025 & 4.08 & 6.77 & 55.04 \\
                DRPCA-Net~\cite{drpcanet} & 2025 & 1.17 & 73.84 & 38.86 \\
                IRPNet~\cite{irpnet} & 2026 & 32.34 & 26.63 & 50.14 \\
                PQGNet~\cite{pqgnet} & 2026 & 1.19 & 9.89 & 27.30 \\
                FGARNet~\cite{fgarnet} & 2026 & 8.40 & 11.80 & 53.95 \\
                \hline
                SAIST~\cite{saist} & 2025 & 389.57 & - & - \\
                \cellcolor[HTML]{E0E9F7}{DGNet(Ours)} & \cellcolor[HTML]{E0E9F7}{} & \cellcolor[HTML]{E0E9F7}{5.34} & \cellcolor[HTML]{E0E9F7}{8.06} & \cellcolor[HTML]{E0E9F7}{75.61} \\
                \hline
		\end{tabular}
	\label{Computational_Efficiency}
	\vspace{-15pt}
	\end{center}
\end{table} 

\subsection{Computational Efficiency}
During the training stage, DGNet introduces CLIP for feature modulation and alignment. It utilizes fixed prior knowledge to modulate visual features and leverages fixed consensus knowledge to guide the learning direction. However, during inference, the CLIP text/image encoder is not involved, and thus no additional computational overhead is introduced.
Therefore, DGNet maintains a relatively efficient inference time.
We evaluate the computational complexity of the models using the number of parameters (Params), floating-point operations (FLOPs), and frames per second (FPS). As shown in Table \ref{Computational_Efficiency}, DGNet achieves a high inference speed of 75.61 FPS while maintaining a relatively low parameter count (5.34M) and FLOPs (8.06G), demonstrating strong competitiveness among existing SOTA methods.
Compared with computationally intensive models such as SAIST, SCTransNet, PBT, and IRPNet, 
DGNet significantly reduces computational cost while still achieving SOTA performance. 
Overall, the proposed DGNet not only delivers superior detection performance but also maintains an efficient and reasonable computational complexity.

\section{Conclusion}
\label{sec:conclusion}
In this work, we investigate the IRSTD task and identify two key challenges in existing text-guided methods: semantic entanglement caused by single specific text and deployment limitations introduced by image-specific prompts. To address these issues, we propose a novel Dual-knowledge Guided Network (DGNet) based on multiple generalizable texts. Specifically, we first design a PWM module, which leverages dual textual priors to precisely disentangle entangled semantics in the frequency domain. Next, we propose a CDA loss, which constrains the optimization process along a cross-sample consensus trajectory, forming a directed alignment path from `complex background' to `salient target', eliminating the reliance on external large models during inference. Extensive experiments on three public datasets demonstrate the effectiveness and superiority of the proposed DGNet and the designed loss function.

\clearpage\section{Acknowledgments}

This work was supported in part by the National Natural Science Foundation of China (NSFC) under Grant 62576194, in part by the ``Key R\&D Program of Shandong Province, China'' under Grant 2025CXGC020101, and in part by the project Youth Science Fund (B) supported by Shandong Provincial Natural Science Foundation under Grant ZR2026QB12.

\bibliographystyle{ACM-Reference-Format}
\balance
\bibliography{DGNet}


\end{document}